%% file: neurips_2026.tex
\documentclass{article}

\usepackage[preprint]{neurips_2026}

\usepackage[utf8]{inputenc} % allow utf-8 input
\usepackage[T1]{fontenc}    % use 8-bit T1 fonts
\usepackage{hyperref}       % hyperlinks
\usepackage{url}            % simple URL typesetting
\usepackage{booktabs}       % professional-quality tables
\usepackage{amsfonts}       % blackboard math symbols
\usepackage{nicefrac}       % compact symbols for 1/2, etc.
\usepackage{microtype}      % microtypography
\usepackage{xcolor}         % colors

\usepackage{amsmath}        % blackboard math symbols
\usepackage{cleveref}
\crefname{section}{Section}{Sections}
\Crefname{section}{Section}{Sections}

\usepackage{subcaption}
\usepackage{graphicx}
\usepackage{tikz}
\usepackage{xspace}
\usepackage{wrapfig}
\usepackage{enumitem}

\newcommand{\X}{\textcolor{black}{FG-Attn}\xspace}

\newcommand*\circled[1]{\tikz[baseline=(char.base)]{
            \node[shape=circle,fill,inner sep=1pt] (char) {\textcolor{white}{#1}};}}

\title{\X: Leveraging Fine-Grained Sparse Attention in Video Diffusion Models}

\author{%
  Sankeerth Durvasula$^{1,2,3}$ \And
  Kavya Sreedhar$^{4}$ \And
  Zain Moustafa$^{1}$ \And
  Suraj Kothawade$^{4}$ \AND
  Tianlei Pang$^{1}$ \And
  Ashish Gondimalla$^{5}$ \And
  Suvinay Subramanian$^{4}$ \AND
  Narges Shahidi$^{*,4}$ \And
  Nandita Vijaykumar$^{*,1,2}$
}

\begin{document}

\maketitle

% \footnotetext{$^{*}$Joint mentorship. $^{1}$Department of Computer Science, University of Toronto, Toronto, Canada. $^{2}$Vector Institute, Toronto, Canada. $^{3}$Sankeerth Durvasula was supported by an internship at Google. $^{4}$Google, Mountain View, USA. $^{5}$Google, Sunnyvale, USA.}

\renewcommand{\thefootnote}{}
\footnotetext{$^{1}$Department of Computer Science, University of Toronto, Canada. $^{2}$Vector Institute, Toronto, Canada. $^{3}$Sankeerth Durvasula was supported by an internship at Google. $^{4}$Google, Mountain View, USA. $^{5}$Google, Sunnyvale, USA. $^{*}$Joint mentorship. Correspondence to: Sankeerth Durvasula \texttt{sankeerth@cs.toronto.edu}}
\renewcommand{\thefootnote}{\arabic{footnote}}

\begin{abstract}
\input{sections/abstract}

\end{abstract}

\input{sections/introduction5}

\input{sections/background}

\input{sections/related_work}

\input{sections/analysis}

\input{sections/method2}
\input{sections/results}

\input{sections/conclusion}

% \begin{ack}
% \input{sections/ack}
% \end{ack}

% \section*{References}
\bibliographystyle{abbrvnat}
\bibliography{refs}

%%%%%%%%%%%%%%%%%%%%%%%%%%%%%%%%%%%%%%%%%%%%%%%%%%%%%%%%%%%%
\appendix

\include{sections/appendix}

\end{document}

%% file: sections/abstract.tex
Using diffusion transformers for media generation may require evaluating attention over extremely long sequences, with attention layers accounting for the majority of generation latency. Exploiting sparsity in attention maps offers a promising opportunity to reduce this cost. 
In this work, we show that attention maps in diffusion transformers exhibit significant fine-grained sparsity in video generation models.
Existing sparse attention methods, however, are too coarse-grained, leaving a large fraction of redundant computation unaddressed, or incur high overheads at finer granularity.  
% Leveraging this efficiently on modern GPUs is challenging, as fine-grained skipping introduces irregular memory access and can reduce tensor core utilization.
We propose \X, a novel, low-overhead fine-grained sparse attention mechanism that skips score computations at the granularity of a $M\times N$ tile, where $N\ge1$ and $M\ge16$, and where each block is the result of query-key dot products between $M$ queries and $N$ keys. \X addresses the key challenge of hardware underutilization in sparse attention kernels on GPUs, without incurring the overheads of irregular memory access and redundant operations.
\X can fully supersede existing sparse attention methods and extend block sparse attention methods to finer granularities on modern GPUs. 
% We demonstrate that \X achieves up to $85\%$ runtime of Flash Attention 3 ($81\%$ on average) on Hopper GPUs. 
At 70\% sparsity, \X is up to $2.45\times$ faster than the state-of-art FlashInfer, and reduces attention kernel time by 14.7\% on average. \X speeds up end-to-end video generation times by up to $1.40\times$ ($1.18\times$ on average) over Flash Attention 3.

% We introduce an efficient asynchronous gather-load primitive that loads only the sparse set of key/value vectors into tensor core-compatible tiles in the on-chip GPU shared memory, hiding the overhead of irregular memory access. We then develop a persistent load balancer to distribute attention computation across thread blocks evenly, ensuring efficient GPU utilization.
% \todo{what number do I report!!}.
% \todo{X; update the number again} speedup on the attention for state-of-art video models on an H100 GPU.
% \todo{can do 128x1, mask scheme, but with inefficiencies. for differnet sizes.  FI comparison for 32x32. }

%% file: sections/Introduction5.tex
\section{Introduction}
\label{sec:introduction}

Diffusion transformers~\citep{dit} (DiTs) are widely used for generative modeling because they can capture complex, high-dimensional data distributions over real-world data. This capability makes them particularly effective for media generation tasks spanning videos~\citep{wan, hunyuanvideo, ltx2}, audio~\citep{diffwave, ltx2}, 3D models~\citep{xiang2024structured, xiang2025trellis2}, and images~\citep{flux, flux_kontext, sd3}. 
% When trained on large-scale datasets, these models can synthesize realistic, high-quality content, enabling transformative applications such as advanced video editing and 3D modeling. 
DiTs generate synthetic data by iteratively refining a latent-space representation.
For example, in video generation, each video is represented as a long sequence of embeddings that are progressively refined by a DiT before being decoded into a video.
% , as illustrated in Fig.~\ref{fig:dit_generation}. 
% A noisy video is first encoded into a set of embedding vectors (the \emph{latent-space} representation) using a variational autoencoder (Fig.~\ref{fig:dit_generation_a}). 
% These embedding vectors are flattened into a single sequence. 
% A DiT progressively refines the embeddings to produce clean video-frame embeddings (Fig.~\ref{fig:dit_generation_b}).
Even short, low-resolution videos yield extremely long embedding sequences. For example, Wan 2.1 1.3B~\citep{wan} encodes a 5-second 720p video into $\sim75000$ embedding vectors, requiring 5 minutes to generate on an H100 GPU; Wan 2.1 14B requires over 15 minutes. The majority of this generation time is incurred by the transformer's attention layers. Since attention scales quadratically with sequence length, higher resolution or longer videos rapidly increase latency: generating a 10-second video requires twice as many embeddings and roughly $4\times$ attention computation.
% \sd{Fig.~\ref{fig:attnbreakdown} breaks down the runtime by operations. ``Others'' includes operations such as encoding text tokens and initial noisy frames into the latent embedding space.}
For Wan 2.1 1.3B, attention layers take $61\%$ of the overall time to produce 49 frames and $91\%$ for 81 frames with Flash Attention 2~\cite{fa2}.

It is well known that the attention computation in DiTs contains significant redundancy since many query–key pairs yield negligible attention scores~\cite{svg, sparge}. Skipping computation of these negligible scores can significantly accelerate video generation. \emph{Sparse attention} mechanisms that enable this require (1) determining the sparse attention \emph{mask}, i.e., which subset of the attention scores can be skipped, and (2) sparse attention kernel that can efficiently skip attention computation without incurring implementation overheads. Prior works propose a range of techniques for (1), e.g., ~\cite{svg, svg2}. For (2), all existing sparse attention mechanisms use \emph{block-sparse attention} kernels. These block-sparse attention mechanisms~\cite{flexattention, bsa} divide the attention score matrix into \emph{tiles} of size M×N, for $M$ query tokens and $N$ key tokens--referred to as a \emph{block}. Computing a block of attention scores is skipped only if \emph{all} scores in that block are predicted to be near zero. FlashInfer~\cite{flashinfer} enables \emph{variable} block sparse attention, where different regions of the attention map can use different block sizes. Thus, FlashInfer can skip computation over irregularly sized blocks rather than requiring a single fixed block size.

We observe that attention maps in diffusion transformers exhibit \emph{fine-grained} sparsity: many query–key products are near zero even when others in the same block are not. Exploiting this finer structure has the potential to substantially reduce computation (floating-point operations, or FLOPs). For example, skipping $16\times16$ blocks of attention can reduce the required FLOPs by up to $70\%$, compared to only $\sim15\%$ with $64\times64$ block sparsity, without noticeable degradation in video quality (\cref{sec:analysis_skip_attn_fine_granularity}).  However, all existing block-sparse attention implementations (except FlashInfer~\cite{flashinfer}, which is discussed below) exploit sparsity only at a coarse block granularity.
They skip attention computation over large contiguous regions of the attention score matrix, typically blocks of size $128 \times 128$. Reducing the block size in these implementations is not directly supported.
The reason for this is that high-performance block-sparse attention is implemented using dense \emph{tile} operations: to compute $M$ output vectors, the kernel loads $M$ query vectors (referred to as a \emph{tile} of queries) and iterates through the key/value sequence by loading $N$ key/value vectors (\emph{tiles} of keys/values) at a time, from main memory into the GPU's shared memory. Block-sparse attention either computes all corresponding scores in an $M \times N$ block or skips it. The tile dimensions M and N are typically set to 128 to match hardware requirements, such as tensor core sizes and memory transaction widths. Reducing them below these sizes degrades hardware utilization, making fine-grained block-sparse attention at, say, $16\times 16$ inefficient.
A consequence of using coarse-grained block-sparse attention is that prior sparse-video methods rely on token clustering~\cite{svg2, luo2026training} to induce a block-sparse structure before attention is evaluated. This clustering process has to be performed online at inference time, thereby adding significant overhead. Larger clusters may alleviate this overhead, but compromise generation accuracy.
% FlashInfer introduces variable sparse attention that supports arbitrary block sizes.

% A consequence of using coarse-grained block-sparse attention is that prior sparse-video methods rely on token clustering~\cite{svg2, luo2026training} to induce a block-sparse structure before attention is evaluated. This creates two problems: 1) it incurs significant overhead in clustering and reordering query/key tokens during inference, and 2) it loses accuracy at fewer clusters, while more clusters preserve accuracy but reduce the speedup.
% \todo{this creates two problems.. incurs significant overhead...., sacrifices accuracy...}

%\textbf{Our goal} in this work is to enable highly efficient \emph{fine-grained block-sparse attention} that skips blocks at a granularity of \todo{M contiguous queries, N contiguous keys.} $M\times N$, with $M\ge16$, $N\ge1$ on modern GPUs. 
%To this end, we propose X, that comprises a highly efficient fine-grain sparse attention kernel and sparse mask determination strategy for fine-grain blocks. There are two major challenges in avoiding significant efficiency overheads when enabling finer-grained block sizes: 

% Thus skipping
\textbf{Our goal} in this work is to enable highly efficient \emph{fine-grained} block-sparse attention that skips blocks at a granularity of $M\times N$, with $M\ge16$, $N\ge1$ (corresponding to $M$ contiguous queries and $N$ contiguous keys) on modern GPUs.
% corresponding to $M$ contiguous queries and $N$ contiguous keys 
To this end, we propose \X, which comprises a highly efficient fine-grain \emph{sparse attention kernel implementation} and a \emph{sparse mask determination strategy} to identify fine-grain blocks to be skipped. There are two major challenges in avoiding significant efficiency overheads when enabling finer-grained block sizes: 

\textbf{Challenge 1.} Naively reducing the $N$ dimension to, say, $N=16$ or $N=1$, means that at each iteration over the key/value sequence, the kernel loads only a small number keys to multiply with $M$ query tokens. This is inefficient because high-performance attention kernels use larger tiles, typically $N=128$, to amortize instruction scheduling, synchronization, and memory-access latencies, while keeping tensor cores well utilized.
% Naively reducing the $N$ dimension to, say, $N=16$ or $N=1$, means that at each iteration over the key/value sequence, the kernel loads only a small number keys to multiply with $M$ query tokens. This is inefficient because high-performance attention kernels use larger tiles, typically $N=128$, to amortize instruction scheduling, synchronization, and memory-access latencies, while keeping tensor cores well utilized.
FlashInfer addresses this by gathering a sparse set of $N=128$ key/value vectors at each iteration and packing them into a tile in shared memory. However, this gather step introduces indirect memory accesses, per-token address computation, and scheduling overhead, which can dominate runtime at very fine sparsity granularities, as we demonstrate.

\textbf{Challenge 2.} Reducing the $M$ dimension (which maps to M queries) to smaller sizes than 128 leads to the same challenge where the hardware becomes underutilized. Gathering sparse queries into a larger block size to improve utilization would be significantly inefficient, because now queries would have to be loaded for each tiled multiply. In the baseline implementation, the M queries need only be loaded \emph{once} initially and are then multiplied over all keys iteratively.

\textbf{\X: Fine-grain sparse attention kernel.}
%, an efficient fine-grained sparse attention mechanism that skips computing scores corresponding to fine-grain blocks of multiples of $M \times N$, $M\ge16, N\ge 1$ of the attention map, i.e., attention scores produced by a group of M=16 contiguous queries and one key.
To efficiently gather key/value tokens (\textbf{challenge 1}), \X implements an asynchronous gather-load mechanism to fetch sparse key/value vectors from irregular memory locations and pack them into tiles of size $N=128$ in shared memory. To hide the high latency of these indirect loads, the gather-load is \emph{parallelized} and \emph{pipelined} with attention computation (\cref{sec:gather_load}). 
To support small query tile sizes $M\le128$ (\textbf{challenge 2}), \X groups together smaller query tiles (for example, eight 16-query tiles into one 128-query tile) before iterating through keys and values. This grouping is based on the sparse mask, so query tiles that attend to similar key/value sets are grouped together to reduce redundant key/value loads while enabling efficient hardware utilization (\cref{sec:querygroup_merging}). After merging, different output tiles, each computed from a query tile of size $M$, may attend to different numbers of key/value tokens and therefore require different amounts of work. Tiles that iterate over fewer tokens may finish earlier than tiles that iterate over more tokens, leading to hardware underutilization from \emph{load imbalance} across the GPU Streaming Multiprocessors (SMs). \X introduces a scheduling mechanism in the attention kernel that dynamically partitions work equitably across SMs to reduce idleness (\cref{sec:querygroup_merging}).

% FlashInfer uses a longest-processing-time scheduling policy, but this can introduce additional scheduling overhead.
% To address \textbf{Challenge 3}, \X uses a \emph{persistent} GPU-side scheduler that assigns output tiles to SMs in longest-processing-time-first order, based on the number of key/value tokens each tile attends to. Since scheduling is fused into the attention kernel, \X avoids CPU-side scheduling overhead.

% 
\textbf{\X: Mask determination.} To exploit fine-grained sparsity, \X must determine which $M \times N$ query--key blocks can be skipped without computing the full attention matrix. \X uses a training-free strategy for this purpose: For each query tile and key tile, we compute their mean query and mean key vectors. For fine-grained masks, these mean vectors provide effective tile-level summaries for estimating the attention score of the corresponding block. We then retain only the blocks selected by top-$p$ masking~\cite{holtzman2019curious} (\cref{sec:method_maskdetermination}).

% \X proposes a training-free strategy to identify which key tile to load for a given query tile: we compute attention using the mean query and mean key for each tile, and retain only the top-\(p\) fraction of blocks by score~\cite{holtzman2019curious} (see~\cref{sec:method_maskdetermination}).

\vspace{-0.1cm}

We demonstrate the efficiency of \X in two ways: First, we use a microbenchmark to demonstrate that our kernel is faster than all existing sparse attention kernels for all sparsity levels. At a sparsity of 70\%, \X speeds up attention computation by up to $2.45\times$ in comparison to the state-of-art FlashInfer. Even without accounting for the planning time overhead, \X reduces the attention kernel latency by 14.7\% on average. Second, for end-to-end video generation using state-of-art generation models such as HunyuanVideo~\cite{hunyuanvideo}, Wan 2.1~\cite{wan}, and LTX-2~\cite{ltx2}, \X provides speedups of up to $1.40\times$ ($1.18$ on average), compared to Flash Attention 3, with an average PSNR of 25.1 relative to dense attention. 

% We demonstrate how \X can be applied to video generation models to enable faster generation times without significantly sacrificing quality. Compared to using FlashInfer sparse attention~\cite{flashinfer}, \X achieves a higher speedup of up to $1.52\times$ ($1.46\times$ on average) on microbenchmarking. Our contributions are:

\vspace{-0.1cm}

\begin{itemize}[leftmargin=*,labelsep=0pt,itemsep=0.5pt]
\item To our knowledge, \X is the fastest existing sparse attention kernel implementation on modern GPUs. Thus, \X's fine-grain sparse kernel implementation can flexibly supersede all existing block sparse attention mechanisms. 
\item We demonstrate that video diffusion models contain substantial fine-grained sparsity in their attention maps. We show that existing block-sparse attention mechanisms do not fully leverage this sparsity. We identify and address the key challenges towards a practical and efficient fine-grain sparse kernel implementation on GPUs.
% We demonstrate that video diffusion models contain a significant amount of fine-grain sparsity in their attention maps that are not leveraged by existing block-sparse attention methods. \todo{merged with prev., fix}
% \item We demonstrate that sparsity patterns remain stable across denoising iterations, enabling a cache-based thresholding strategy that avoids recomputation while preserving accuracy.
\item We propose a lightweight strategy for sparse mask generation that operates entirely without retraining. This enables seamless integration with state-of-the-art video DiTs.
% \item We show that our sliced attention mechanism can fully supersede existing block-sparse attention methods in DiTs, achieving performance on par with or better than all prior coarse-grained approaches while incurring only a small loss in accuracy.
\end{itemize}

\vspace{-0.25cm}

%% file: sections/background.tex
\section{Background}
\label{sec:background}
% \subsection{} 
\label{sec:background_attnimpl}
% \vspace{-0.2cm}

% \textbf{GPU Architecture.} \todo{} ~\cref{sec:background_gpu} of the Appendix.
\textbf{Modern GPU Architecture.} A GPU consists of many streaming multiprocessors (SMs), each running thousands of threads organized into thread blocks and 32-thread warps; each SM has a small, fast shared memory backed by larger but slower high-bandwidth memory (HBM). We refer the reader to~\cref{sec:background_gpu} of the Appendix for a more detailed description of these components.

\begin{figure}[!htb]
\vspace{-0.3cm}
\centering
\begin{subfigure}{0.47\textwidth}
    \includegraphics[width=\linewidth]{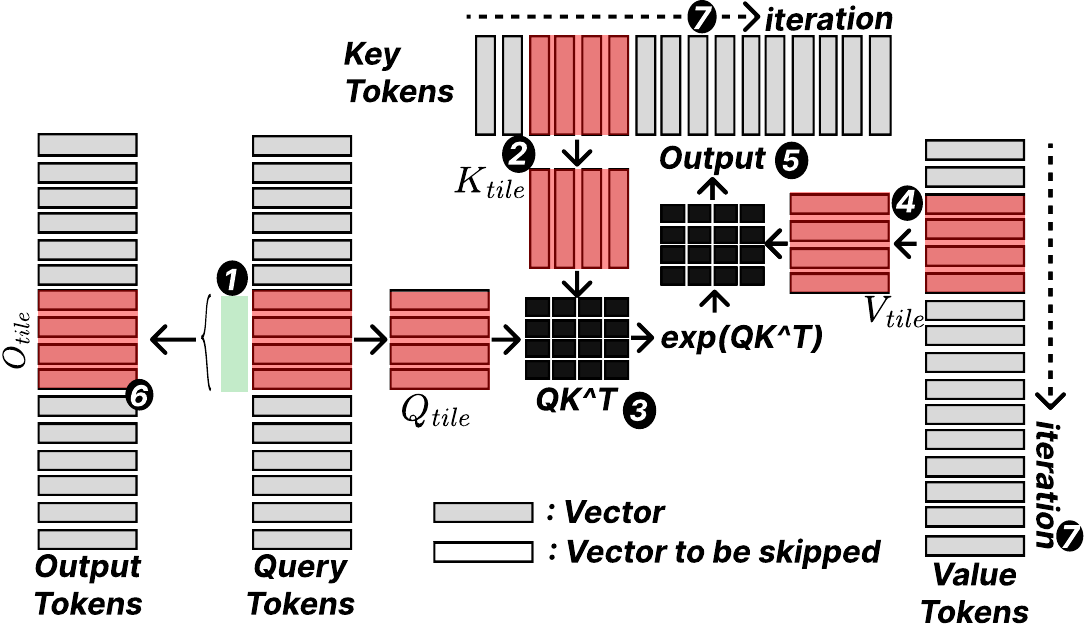}
    \caption{Flash Attention: tiles of keys, values are loaded into shared memory. Tensor cores are used to compute a partial sum of the output vector.}
    \label{fig:flashattention}
\end{subfigure}\hfill
\begin{subfigure}{0.47\textwidth}
    \centering
    \includegraphics[width=0.76\linewidth]{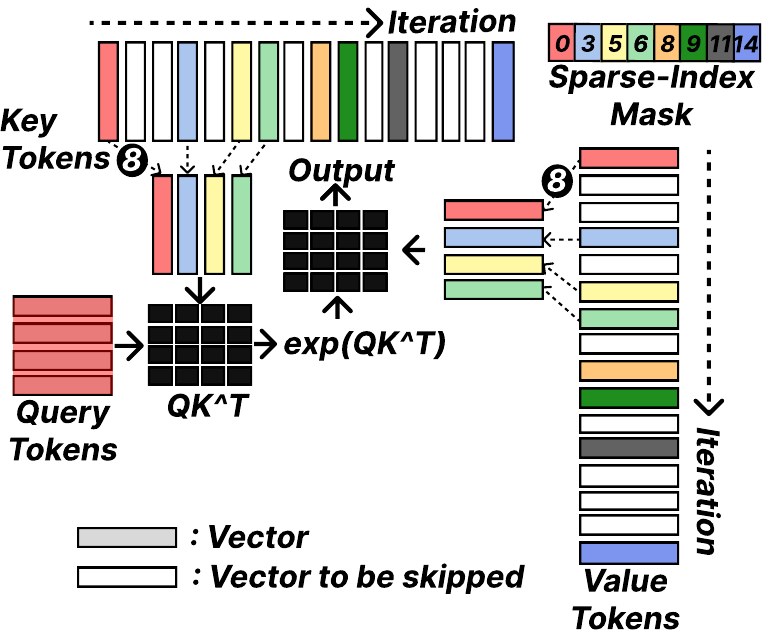}
    \caption{\X: only relevant keys and values corresponding to indices in the sparse index mask are loaded into shared memory.}
    \label{fig:method_implementationverview}
\end{subfigure}
\vspace{-0.1cm}
\caption{Implementation overview of Flash Attention~\citep{fa} and \X.}
\vspace{-0.4cm}
\end{figure}

\textbf{Efficient dense attention.} Efficient attention implementations (Flash Attention 1, 2, and 3~\citep{fa, fa2, fa3}) fuse the attention score computation $\mathbf{P}=\mathbf{softmax}\left( \mathbf{QK}^\top/\sqrt{D}\right)$ and multiplication of attention scores with values $\mathbf{O}=\mathbf{PV}$ for fast execution. 
Fig.~\ref{fig:flashattention} depicts how Flash Attention is implemented in a GPU. The kernel takes queries, keys, and values ($\mathbf{Q}, \mathbf{K}, \mathbf{V}$) matrices as input and produces an output matrix $\mathbf{O}$. The queries, keys, values, and output tokens form a sequence of $N$ tokens (gray bars), as shown in the figure. 
% A set of contiguous output tokens for a particular head is computed within a block. 

GPU threads are organized into thread blocks, where each thread block cooperatively computes a tile of the output for a particular attention head.
In the figure, the output tokens labeled $\mathbf{O}_{\mathrm{tile}}$ highlighted in red is computed by the threads of one thread block. This corresponds to a set of query tokens highlighted in green. To compute $\mathbf{O}_{\mathrm{tile}}$, the block first loads a corresponding set of queries $\mathbf{Q}_{\mathrm{tile}}$ into the GPU's scratchpad (shared) memory~\circled{1}. Then, the first set of key and value tokens, $\mathbf{K}_{\mathrm{tile}}$~\circled{2} and $\mathbf{V}_{\mathrm{tile}}$~\circled{4}, is loaded into shared memory. 
This tile is used to compute $\mathbf{Q}_{\mathrm{tile}}\mathbf{K}_{\mathrm{tile}}^T$ using the tensor core~\circled{3}, followed by exponentiation to compute $e^{\mathbf{Q}_{\mathrm{tile}}\mathbf{K}_{\mathrm{tile}}^T}$. This result stored in registers is then multiplied by the corresponding $\mathbf{V}_{\mathrm{tile}}$ using the tensor core~\circled{5}. The result of the computation is added to $\mathbf{O}_{reg}$, a slice of output in registers~\circled{6}. In the next iteration, the next tile $\mathbf{K}_{\mathrm{tile}}, \mathbf{V}_{\mathrm{tile}}$ in the sequence is loaded into shared memory, and computation is repeated~\circled{7}. 
The thread block also keeps the running sum of exponents, $e^{\mathbf{Q}_{\mathrm{tile}}\mathbf{K}_{\mathrm{tile}}^T}$, required for softmax normalization in registers. The running sum is used to normalize the accumulated partial output before writing the final $\mathbf{O}_{\mathrm{tile}}$.
% the running sum is not shown in the figure.
% The thread block accumulates the running softmax normalization factor in the registers

\textbf{Efficient block sparse attention.}
\label{sec:background_blocksparseattention}
Block-sparse attention follows the same tiled execution pattern as dense Flash Attention, but uses a block-level sparse mask to decide which key/value tiles are processed for each query tile. For a given $\mathbf{Q}_{\mathrm{tile}}$, if the mask selects a key/value block, the corresponding $\mathbf{K}_{\mathrm{tile}}$~\circled{2} and $\mathbf{V}_{\mathrm{tile}}$~\circled{4} are loaded into shared memory and used in the tensor core attention computation. Otherwise, that key/value block is skipped entirely: it is not loaded into shared memory, and the corresponding attention scores and values are not computed. This reduces both, expensive memory traffic and computation in proportion to the number of skipped key/value blocks.

% The sum over the exponents $e^{\mathbf{Q}_{\mathrm{tile}}\mathbf{K}_{\mathrm{tile}}^T}$ is also computed in registers to hold the denominator of the softmax function (not shown in the figure).

\vspace{-0.2cm}

%% file: sections/related_work.tex
\section{Related Work}
\label{sec:related_work}
% \vspace{-0.2cm}

\textbf{Block-sparse attention.}  
Several implementations of block-sparse attention~\citep{bsa, fa, flexattention, flashmask} 
propose coarse-grained mechanisms that skip entire tiles of attention scores, typically at $64\times64$ 
or $128\times128$ granularity in half-precision. These methods have been widely adopted in LLM inference~\citep{minference, nsa}. However, current block-sparse methods do not support smaller tile sizes: reducing block-size either fails to compile or leads to severe hardware underutilization due to tensor core width constraints (\cref{sec:analysis_attentionsparsity}). FlashInfer~\cite{flashinfer} introduced variable block-sparse attention, enabling finer-grained sparsity beyond what earlier block-sparse implementations support. However, FlashInfer incurs nontrivial overhead in the planning stage when scheduling work to SMs at runtime and pads query-key dot product computations at tiles smaller than $128\times1$, resulting in wasted compute. We compare \X against FlashInfer  in~\cref{sec:results_attn_kernel_analysis}.

\textbf{Sparse attention for video diffusion models.} Many prior works~\citep{radial, svg, xu2025xattention} leverages fixed sparsity patterns based on empirical observations of attention scores. SpargeAttention~\citep{sparge} predicts a mask based on pooling blocks of queries and keys. These methods use block-sparse attention to compute attention and are orthogonal to our proposed approach. They can leverage \X for their underlying sparse attention mechanism to exploit fine-grain sparsity.
Clustering-based methods~\citep{svg2, luo2026training} reorder similar sets of keys and queries together in order to promote block sparsity in attention score computation. While clustering tokens is a well-known technique to promote sparsity to use block-sparse attention~\cite{vyas2020fast}, these methods incur high costs in clustering similar query and key tokens. In contrast, \X directly leverages fine-grained masks without requiring global token clustering or reordering to expose coarse block structure.
Training-based methods~\citep{vsa, zhang2025fast, zhang2026spargeattention2, vmoba} learn to predict the sparsity mask. All these methods, however, are limited to coarse-grained block skipping. For example, Video Sparse Attention~\cite{vsa} can also leverage \X for its sparse attention mechanism, and are thus orthogonal and complementary. 
% We adopt the mask determination method proposed by SparseVideoGen2~\cite{svg2} for \X~\cref{sec:results} to evaluate sparse video generation.

\textbf{Other techniques to accelerate video diffusion.}
SageAttention~\citep{sageattention, sageattention2, sageattention2++, zhang2025sageattention3} 
uses quantization to speed up attention layers in transformers. Since quantization and sparsity are orthogonal, \X can be applied on top of quantization-based approaches. 
Caching-based approaches such as DeepCache~\cite{deepcache}, TeaCache~\cite{teacache}, and TaoCache~\citep{fan2025taocache}  exploit temporal redundancy across denoising steps. 
% Distillation-based methods fewer denoising iterations
These approaches are orthogonal to our work and can be combined with \X to further accelerate attention computation. 

% \vspace{-0.2cm}

%% file: sections/analysis.tex
% \vspace{-0.4cm}
\section{Fine-Grained Sparsity in Attention Computation}
\label{sec:analysis}

\label{sec:analysis_videodit_modelperformance}

% \vspace{-0.3cm}

% Sparse attention reduces the quadratic cost of self-attention by computing only a subset of the entries in the attention matrix.
% Given a sequence of $N$ queries and $N$ keys, dense attention computes the matrix $QK^\top \in \mathbb{R}^{N \times N}$.
% Sparse attention reduces this computational cost by computing only a selected subset of elements in $QK^\top$ that correspond to significant attention scores.
% Dense attention therefore scales poorly with sequence length: its complexity is $O(N^2)$, whereas feed-forward layers scale as $O(N)$.

% Long sequence lengths are common in generative models such as video diffusion models.
% In video diffusion transformers, a video is represented as a sequence of latent embeddings, with each embedding corresponding to a local spatiotemporal region.
% A five-second video at $480 \times 832$ resolution can produce approximately $N=32{,}000$ latent embeddings (in Wan2.1~\cite{wan}).
% Processing attention over such long sequences requires massive computation having complexity $O(N^2)$, making dense attention the dominant cost even for short videos. 

\begin{figure}[!htb]
% \vspace{-0.2cm}
\centering
\begin{subfigure}{0.45\textwidth}
    \centering
    \includegraphics[width=0.76\linewidth]{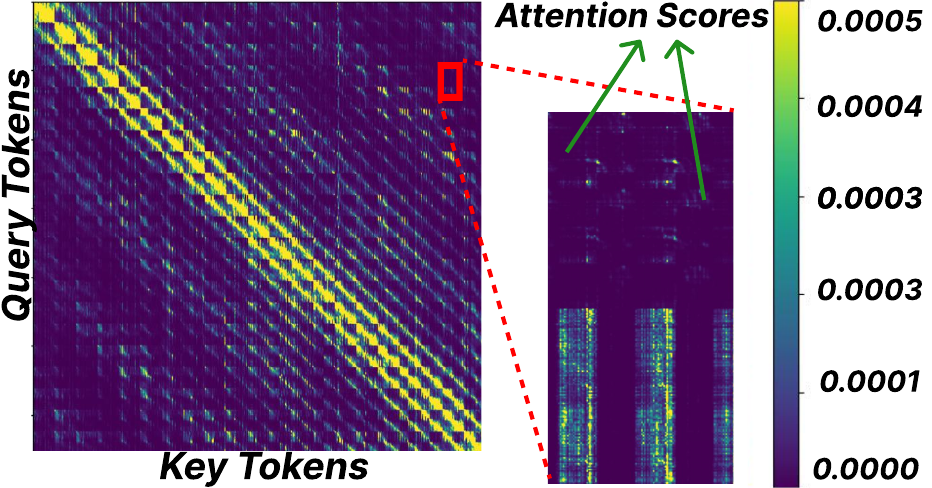}
    \caption{Sparsity in attention computation: attention scores are highly sparse and irregular.}
    \label{fig:distribution_of_attnvalues}
\end{subfigure}~
\begin{subfigure}{0.45\textwidth}
    \centering
    \input{tables/blocksparse}

    \caption{Sparsity vs. block size: \% of $M\times M$ attention map blocks with all scores $\le$ threshold.}
    \label{tab:attn_block_sparsity}      
\end{subfigure}    
\vspace{-0.2cm}
\caption{Attention maps are very sparse. Block sparsity increases significantly at finer granularity.}
\vspace{-0.4cm}
\end{figure}

\label{sec:analysis_attentionsparsity} \label{sec:analysis_skip_attn_fine_granularity}
Fig.~\ref{fig:distribution_of_attnvalues} shows a heatmap of the attention scores in one attention head of the Wan 2.1 1.3B vDIT model.
From the figure, we observe that the vast majority of attention scores are close to zero. Over $85\%$ of the scores are below $0.5/N$ (N is the sequence length) in this example. Attention computation can be significantly accelerated by skipping these score computations without sacrificing output quality. For sequence length $N$ and model dimension $D$, dense attention costs $O(N^2 D)$ floating-point operations. If only a fraction $\rho \in [0,1]$ of the $N^2$ query–key pairs are retained (i.e., $\rho$ is the nonzero density, $\rho = 1 - \text{sparsity}$), the cost becomes $O(\rho N^2 D)$.
% Sparse attention mechanisms ~\cite{flexattention, bsa, fa}, BlockSparse~\cite{bsa}, 
% and FlashAttention~\cite{fa} accelerate attention by avoiding computing insignificant attention scores.
% We refer to the block size $M$ as the granularity at which the score matrix is partitioned: an $M \times M$ block contains attention scores between $M$ contiguous queries and $M$ contiguous keys.
% Skipping a block, therefore, avoids computing the corresponding $M^2$ query--key dot products.
% To preserve accuracy, however, a block can be skipped only when all scores within the block are negligible. 
Table~\ref{tab:attn_block_sparsity} reports attention map sparsity at different block sizes, 
measured as the fraction of \(M \times M\) blocks with all scores below a threshold of \(0.5/N\), 
where \(N\) is the sequence length. 
Finer blocks (\(16 \times 16\)) yield about 70\% sparsity, while coarser blocks (\(64 \times 64\)) achieve only 22\%. 
This shows that finer granularity offers much greater opportunity for speedup, yet existing block-sparse implementations cannot exploit blocks smaller than \(128 \times 128\).

\vspace{-0.2cm}

% \todo{Explain that Other works do not implement it. - it must be clear by now}

% \begin{wrapfigure}{R}{0.48\textwidth}

% \end{wrapfigure}

% \begin{figure}[!htb] %{L}{0.48\textwidth}
%     \centering
%     \input{tables/blocksparse}
%     \caption{Sparsity vs. block size: \% of $M\times M$ attention map blocks with all scores $\le$ threshold.}
%     \label{tab:attn_block_sparsity}        
% \end{figure}    

%% file: tables/blocksparse.tex
\begin{tabular}{|l|l|l|}
\hline
\textbf{Block size} & \textbf{Sparsity} & \textbf{TFLOPs}\\
\hline
$128\times128$ & 5.5\% & 0.519 \\
$64\times64$ & 22.8\%  & 0.424 \\
$32\times32$ & 47.7\%  & 0.287 \\
$16\times16$  & 70.7\% & 0.161 \\
\hline
\end{tabular}

%% file: sections/method2.tex
\section{Method}
\label{sec:method}
\label{sec:key_challenges_background}
\vspace{-0.2cm}

\subsection{Fetching Keys/Value Tokens Using Gather-Load.}
\label{sec:gather_load}
\vspace{-0.2cm}

In block-sparse attention~\citep{bsa, flexattention}, a block of \(M\) queries \(\mathbf{Q}_{tile}\) and \(N\) contiguous keys \(\mathbf{K}_{tile}\) are loaded into shared memory and multiplied using tensor cores. 
On GPUs, these loads are accelerated by the Tensor Memory Accelerator (TMA). 
In \X, we instead must \emph{gather} \(M\) non-contiguous keys, pack them into a tile \(\mathbf{K}_{rel}\), to compute \(\mathbf{Q}_{tile}\mathbf{K}_{rel}^T\) (\circled{8} in Fig.~\ref{fig:method_implementationverview}). 
To do this efficiently, we introduce a new \emph{gather-load} primitive, which loads sparse key/value vectors from HBM into shared memory (\cref{sec:background_gpu} of appendix) based on the attention mask. 
The corresponding value vectors are loaded with the same indices, and tensor cores compute partial sums across tiles until all relevant slices are processed. Loading sparse key/value vectors requires first computing the addresses of elements specified by the sparse mask before issuing the load.
Efficient gather-load requires (i) fast address generation for sparse indices and (ii) hiding this latency. 
We achieve (i) by parallelizing index-to-address translation across threads in a warp group and (ii) by pipelining (i.e., overlapping) the gather-load with attention computation.

\begin{figure}[!htb]
% \vspace{-0.3cm}
\centering
\begin{subfigure}{0.40\textwidth}
    \centering
    \includegraphics[width=\linewidth]{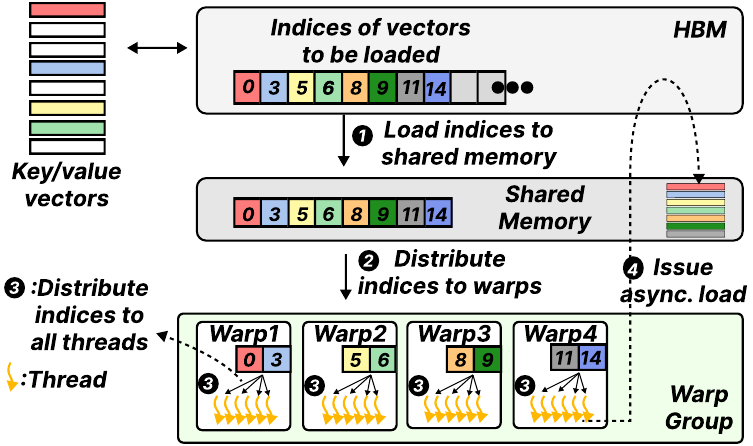}
    \caption{Detailed diagram showing the implementation of the gather + load operation.}
    \label{fig:method_gatherload}   
\end{subfigure}~
 \begin{subfigure}{0.44\textwidth}
    \centering
    \includegraphics[width=\linewidth]{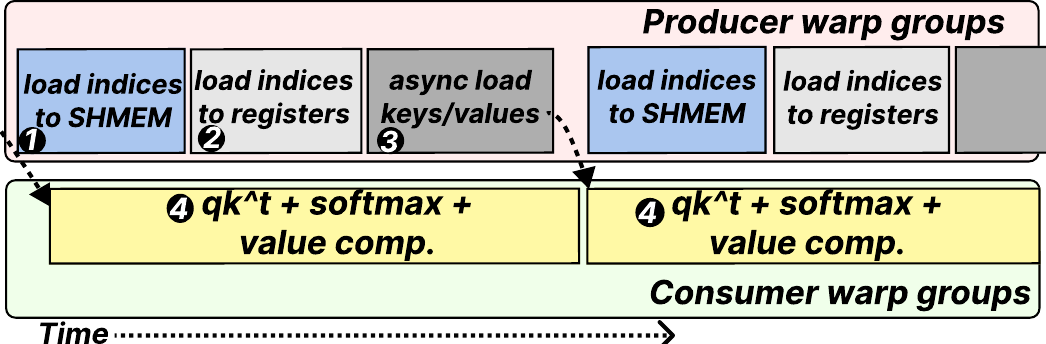}
    \caption{The address generation latency is hidden by attention computation by having the gather-load operation in the producer threads.}
    \label{fig:method_pipelining}
\end{subfigure}
\caption{\X parallelizes gather-load by distributing the selected key indices across threads, which then cooperatively compute addresses and issue loads to fetch the key/value tiles.}
\vspace{-0.5cm}
\end{figure}

\noindent \textbf{The sparse gather-load primitive.} 
% The \emph{gather-load} primitive takes an array of indices, fetches the corresponding key/value vectors from HBM, and assembles them as a contiguous tile in shared memory. 
% \begin{wrapfigure}{R}{0.48\textwidth}
For each group of \(M\) queries, only the relevant keys indicated by the sparse mask are loaded (Fig.~\ref{fig:method_gatherload}). 
Indices are first cooperatively loaded into shared memory~\circled{1} from HBM, then distributed across warps in a warp group~\circled{2}. 
Each warp broadcasts its indices to threads~\circled{3},
% via warp-shuffle, 
after which threads compute addresses and issue asynchronous loads to fetch the sparse vectors into shared memory~\circled{4}.

\textbf{Overlapped indirect key/value load latency with attention computation.}
\label{sec:softwarepipelining}
%Latency of loading keys, values from the sparse index mask can be hidden behind attention computation. 
% On H100 GPUs, 
In modern GPUs, threads in a thread block can be divided into two groups: \emph{producer} threads that load data and \emph{consumer} threads that perform computation. 
Producers load query, key, and value tiles into shared memory, while consumers compute and accumulate partial sums (Fig.~\ref{fig:method_pipelining}). 
At each iteration, the producer loads indices for \(N\) keys from the sparse mask into shared memory~\circled{1}, moves them into registers~\circled{2}, and issues asynchronous loads for the corresponding rows from global memory~\circled{3}. 
The consumer computes attention scores, outputs in parallel~\circled{4}, fully hiding address generation and load latency.

\vspace{-0.3cm}
\subsection{Grouping Query Tiles Attending to Similar Sets of Keys/Values}
\label{sec:querygroup_merging}
\vspace{-0.2cm}

\begin{wrapfigure}{R}{0.43\textwidth}
    \centering
    \includegraphics[width=\linewidth]{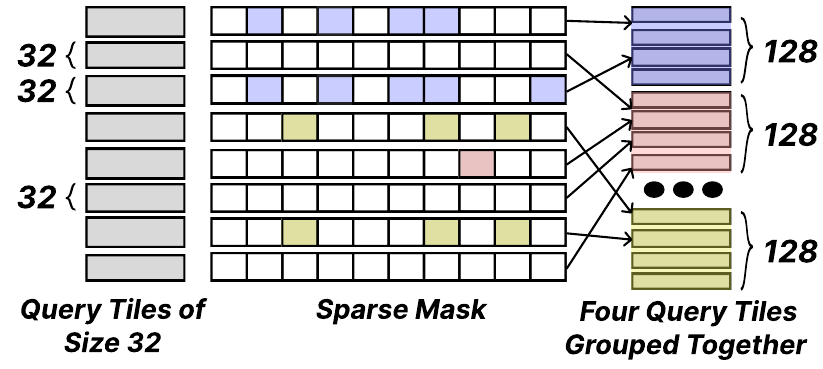}
    \caption{Query tiles (of size 32) attending to the similar sets of keys are grouped together into sets of 128 queries.}
    \label{fig:query_merge}
    \vspace{-0.2cm}
\end{wrapfigure}
% \In Flash Attention 3, threads in each thread block process a fixed block of 128 queries, and iterate over all key tokens to compute attention scores. 
In \X, however, different query tiles (of size $M\le 128$) have to be grouped together to form groups of 128 queries for hardware efficiency (see Challenge 1, \cref{sec:introduction}). These query tiles iterate over different subsets of keys/values, as determined by the sparse mask. When iterating over keys/values, the kernel may need to load redundant key/value tokens. 
For example, in Fig.~\ref{fig:query_merge}, when $M=32$, a single thread block contains four query tiles. If the sparse masks of four consecutive query tiles differ substantially (4 consecutive rows in Fig.~\ref{fig:query_merge}), the merged 128-query tile must load all key/value tokens selected by any of the four tiles. Instead, in \X, we minimize the number of redundant key/value loads. We collect query tiles that load \emph{similar} sets of key/value tiles into groups of tiles containing 128 queries. For example, in Fig.~\ref{fig:query_merge}, the first and third sparse mask rows are similar, indicating that the corresponding query tiles attend to a similar set of key/values. These tiles, along with two other similar tiles, can be grouped into a 128-query set into a thread block.

% disjoint sets of KV tokens, leading to redundant loads. 
To do this, we first define how two query tiles are \emph{similar}: For an $M\times N$ block sparse mask $\mathbf{B}$, we define the distance between two query tiles as:
$
d(i,j)=\sum_k b_{ik} \oplus b_{jk}
$
where $\oplus$ denotes bitwise XOR, and $b_{ik}$ is a boolean. 
This counts the number of key blocks selected by exactly one of the two query blocks. A smaller value, therefore, means that the two query tiles attend to more similar sets of KV tokens. 
For query tile size $M$, each merged group contains $R=128/M$ query tiles. Given the sparse mask $\mathbf{B}$, \X groups query tiles by choosing groups with small distance to a representative tile:
%\vspace{-0.2cm}
\begin{equation}
\label{eq:obj}
\min_{\{G_g,c_g\}}
\sum_g \sum_{i\in G_g} d(i,c_g)
\quad
\text{s.t. } |G_g|=R,\; c_g\in G_g .
\end{equation}
%\vspace{-0.2cm}
Here, $G_g$ is a group of query tiles mapped to one thread block, and $c_g$ is the representative query tile for that group. Minimizing this objective groups query tiles with similar key/value access patterns, reducing redundant key/value loads within each thread block. To find this grouping, we use an approximate, greedy approach: starting from the first ungrouped query tile, we select the remaining $R-1$ ungrouped query tiles with the smallest XOR distance to it, where $R=128/M$. These $R$ tiles form one merged 128-query group assigned to a thread block. We repeat this process until all query tiles are assigned to a group. This heuristic groups query tiles having small distances, reducing the union of key/value tiles loaded by each thread block.
% The block sparse mask for the merged group is the union of the selected masks. Grouping blocks with low XOR distance keeps this union small, reducing redundant KV loads and unnecessary attention computation. After attention is computed, outputs are written back in the original query-token order.

% This merging process is illustrated in Fig.~\ref{fig:query_merge}. Four query blocks of size 32 are merged together into a query group of size 128. The first and third query tiles attending to a similar set of keys indicated in blue are grouped together.}

\textbf{Load balancing.} After query merging, each thread block processes a 128-query tile, but different tiles may attend to different numbers of key/value tokens, as determined by the sparse mask. Since the work for an output tile is proportional to the number of key/value tokens it attends to, different thread blocks can have different runtimes, creating load imbalance across GPU Streaming Multiprocessors (SMs). Prior work addresses this by sorting output tiles on the CPU using a longest-processing-time policy based on their key/value counts, but this CPU-side scheduling creates significant overheads. In contrast, \X performs work selection directly on the GPU inside the persistent attention kernel, allowing thread blocks to fetch new output tiles without CPU-side scheduling. Details are provided in~\cref{sec:load_balance} of the Appendix.

\vspace{-0.2cm}
\subsection{Determining the Fine-grain Sparse Mask with Top-p Masking.}
\vspace{-0.2cm}

\label{sec:mask_determination}
\label{sec:method_maskdetermination}

To exploit fine-grained sparsity, we need a mask that efficiently identifies which query--key slices produce \emph{significant} attention scores. 
We need to do this without computing the full attention score matrix, as this is prohibitively expensive.
% A straightforward approach would recompute this mask at every denoising iteration by evaluating the full attention matrix. For $H$ attention heads and sequence length $S$, this requires computing $H S^2$ scores, which is prohibitively expensive for the long sequences used in video diffusion models and would eliminate much of the benefit of sparse attention.
We note that in the context of video diffusion models, the query and key tokens forming an $M\times N$ attention block correspond to adjacent pixels in space and time. Such adjacent tokens typically exhibit responses similar to those of their surrounding queries and keys.
Motivated by this observation, we propose a simple, lightweight strategy for determining the attention mask. We compute the average of a tile of queries and keys to summarize the corresponding $M\times N$ block of attention scores as follows:
\[
\mathbf{q}^{(i)}_{\mathrm{avg}}=\frac{1}{M}\sum_{m=1}^{M}\mathbf{q}_{i,m},
\qquad
\mathbf{k}^{(j)}_{\mathrm{avg}}=\frac{1}{N}\sum_{n=1}^{N}\mathbf{k}_{j,n},
\qquad
\hat{\mathbf{A}}=
\mathrm{softmax}\!\left(
\mathbf{Q}_{\mathrm{avg}}\mathbf{K}_{\mathrm{avg}}^\top/\sqrt{D}
\right),
\]
where $\mathbf{q}^{(i)}_{\mathrm{avg}}$ and $\mathbf{k}^{(j)}_{\mathrm{avg}}$ are rows of $\mathbf{Q}_{\mathrm{avg}}$ and $\mathbf{K}_{\mathrm{avg}}$, respectively. The block is included if the dot product $\mathbf{q^{(i)}}_\text{avg}\cdot\mathbf{k^{(j)}}_\text{avg}$ is significant. We determine significance by computing the approximate attention distribution between each averaged query block and all averaged key blocks, $\hat{\mathbf{A}}$. containing elements $\hat{a}_{ij}$, $1\le i \le \lceil{S/M\rceil}, 1\le j \le \lceil{S/N}\rceil$. $\hat{a}_{ij}$ denotes the approximate importance of key block $j$ for query block $i$.
For each query block $i$, we sort key blocks by $\hat{a}_{ij}$ and retain the smallest set whose cumulative probability mass is at least $p$. These retained key blocks define the sparse mask used by \X. This mean-query/key top-$p$ mask avoids computing the full token-level attention matrix while preserving the dominant attention regions. Since each approximate score represents an $M\times N$ block, mask construction reduces the number of score computations by roughly a factor of $MN$ compared to dense attention. 

%% file: sections/results.tex
\vspace{-0.2cm}

\section{Results}
\label{sec:results}
\vspace{-0.2cm}
 
We perform two evaluations: First, we micro-benchmark \X against the only other available fine-grained block-sparse attention mechanism to our knowledge, FlashInfer~\cite{flashinfer} (FI), measuring performance on randomly generated masks across various sparsity levels. We evaluate two mask-merging strategies of \X: one obtained by grouping query tiles to minimize objective Eq.~\ref{eq:obj} (\X+merge-XOR, \cref{sec:querygroup_merging}), and one obtained by merging consecutive query tiles (\X+merge-consecutive).
Second, we evaluate video generation times on three open-source video generation models: (1) LTX-2~\cite{ltx2},  (2) Wan 2.1~\cite{wan} 14B, and (3) HunyuanVideo~\cite{hunyuanvideo}. 
We report end-to-end video generation times for the following configurations: 1) the baseline dense Flash Attention 3 (FA3) implementation, 2) FlashInfer, using SparseVideoGen2~\cite{svg2} to generate its sparse mask (FI+SVG2), and 3) \X+merge-XOR with block sizes: $16\times 16$, $32\times 32$, and $64\times 64$ (\X-16, \X-32, \X-64 respectively).
We also present PSNR/SSIM scores for quality comparisons to the baseline video from SVG2, Radial Attention~\cite{radial}, Sparge Attention~\cite{sparge}, and SparseVideoGen~\cite{svg} (SVG). 
Prior works, such as SVG, Sparge, and Radial Attention, also exploit sparse attention, but are built on Flash Attention 2. Thus, we are unable to do an apples-to-apples efficiency comparison and only compare the video quality (PSNR/SSIM). We note, however, that \X is orthogonal to these works: their mask-determination strategies are fully compatible with \X's sparse attention kernel, enabling finer granularity.
% \todo{Given that this is not a fair comparison, the results are shown in the appendix - radial/SVG2}.
% \todo{FG-attn sparse attention enables applying finer granularities with the original works.}
% Thus, FA3 baselines and \X would result in much faster video generation times. For a fair comparison, we only compare the attention speed and generation times with FA3 and SVG2 (which uses Flash Attention 3~\cite{fa3}). We only compare the quality (PSNR/SSIM) and the attention FLOPs reduction with the other prior work.

We implement \X on top of Flash Attention 3~\cite{fa3} in BFloat16 for an H100 GPU. We sample 20 prompts from the Penguin dataset~\cite{hunyuanvideo} for the video, and report video generation times and quality as the arithmetic mean of each metric across all generated videos. 
Similar to prior work~\cite{svg2, svg, radial}, we perform a warmup for the first few denoising iterations during which we compute dense attention for all layers. We set the warmup iterations to $24\%$ of the total for Wan 2.1 and LTX-2, and to $10\%$ for HunyuanVideo.
The remaining denoising iterations are run with sparse attention. We set the top-p threshold to $0.9$.

% \todo{do not re-cite each time}

% \todo{explain the random mask. in order to eval. how efficientlty fg is able to leverage sparisty .. at different attenti onlevels , we generate use a random mask .. with a required amount of sparisty}

% \vspace{-0.3cm}

\subsection{\X Efficiency Analysis}
\label{sec:results_attn_kernel_analysis}
% \vspace{-0.3cm}

\begin{figure}[!htb]
% \vspace{-0.4cm}
\centering
\includegraphics[width=0.95\linewidth]{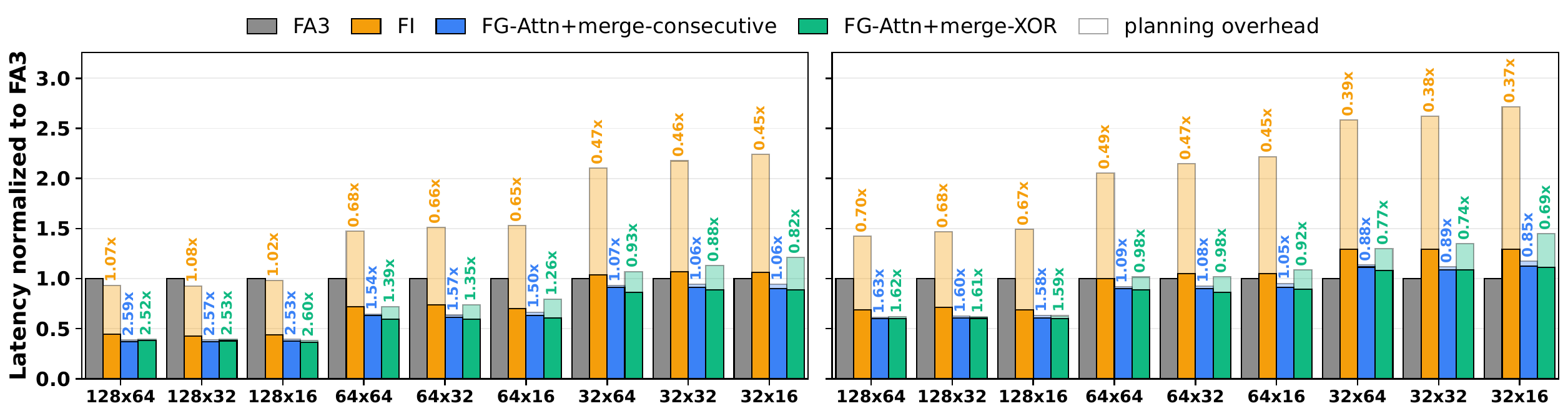}
% \vspace{-0.15cm}
\caption{Attention time at two sparsity levels (70\%, left, and 50\%, right), across block sizes.}
\label{fig:breakdown}
% \vspace{-0.3cm}
\end{figure}

\textbf{\X vs. FI at different sparsities.} To evaluate how efficiently \X computes attention at different levels of sparsity, we generate a random attention mask while controlling the level of sparsity, benchmark the speedup generated, and compare it against FI.
% In this experiment, we aim to compare the speedups generated by \X and FlashInfer~\cite{flashinfer}. \todo{flashinfer in chart: \texttt{FI}.}
Fig.~\ref{fig:breakdown} shows the attention computation latency as a breakdown between planning and kernel computation for different sparse mask block sizes at $70\%$ sparsity and $50\%$ sparsity. From these plots, we make the following observations.
First, at query tile size $128$, \X achieves speedups of up to $2.60\times$ ($2.56\times$ on average) at $70\%$ sparsity, and up to $1.63\times$ ($1.59\times$ on average) at $50\%$ sparsity over FA3. At this tile size, there is no additional overhead from merging masks together, since the query tile is already of size $128$. Thus, the runtime is dominated by the sparse attention kernel, allowing \X to directly benefit from reduced key/value computation.
Compared to FI, \X achieves average speedups of $2.45\times$ and $2.38\times$ at $70\%$ and $50\%$ sparsity, respectively. This is because FI incurs significant overhead from planning and constructing the sparse indices on the CPU for a given mask shape, fully avoided by \X. 
Second, at query tile size $128$, \X reduces kernel computation time relative to FI by $14.7\%$ and $13.8\%$ on average at $70\%$ and $50\%$ sparsity, respectively. This improvement comes from the parallelized gather-load primitive used by \X, as opposed to FI's approach, where individual threads load sparse indices from global memory and then issue the corresponding key/value memory loads.
Third, at smaller query tile sizes ($64\times$ and $32\times$), \X+merge-consecutive and \X+merge-XOR kernels provide smaller speedups: up to $1.57\times$ ($1.54\times$ on average) and up to $1.07\times$ ($1.06\times$ on average) respectively at 70\% sparsity, and a slowdown at 50\% sparsity. This is expected because the random masks used in the microbenchmark exhibit fewer correlations in the key blocks attended by different query blocks. As a result, mask merging has fewer opportunities to exploit overlap, and merge-consecutive and merge-XOR provide similar benefits.
Fourth, at smaller query block sizes $32$ and $64$, the planning time of merge-XOR becomes significant, where it accounts for up to $54.2\%$ and $29.1\%$ of kernel time on average. However, this planning time is much lower than FI's planning time. 
Overall, \X outperforms FI across a full range of sparsity and block sizes. %At query block size $128$, the benefit comes primarily from negligible planning time and faster key/value gather operation. At smaller query block sizes, the benefit is smaller on random masks because there is limited overlap across query blocks, but \X still retains lower planning overhead than FI.

% \textbf{Ablation: \X at different levels of skew, sparsity in sparse mask.}  We evaluate the compute throughput of \X and FI under varying skew in the number of key--value tokens across query blocks. Higher skew leads to load imbalance, which \X mitigates using its persistent work scheduler~\cref{sec:load_balance} of the Appendix. Evaluation details are provided in~\cref{sec:skew_results} of the Appendix. We show that \X outperforms FI across varying degrees of skew and sparsity levels.  

\textbf{Ablation: \X at different levels of skew, sparsity in sparse mask.}  
\label{sec:skew_results}

In this experiment, we aim to evaluate the compute throughput of \X at varying degrees of skew in the number of key--value tokens across query blocks. A greater skew indicates greater variance in the number of KV blocks each query block must iterate over, leading to load imbalance. \X mitigates using its persistent work scheduler (\cref{sec:load_balance} of the Appendix). 
To study this, for each sparse mask, we define the skew as:
$
\mathrm{skew} = \frac{L_{\max} - L_{\min}}{L_{\max}},
$
where $L_{\max}$ and $L_{\min}$ denote the maximum and minimum number of key/value blocks attended by any query tile, respectively. A skew of $0$ corresponds to a balanced mask in which all query tiles attend to the same number of key/value blocks, while larger skew values indicate greater variation in work across query tiles. To generate a mask with a target skew, we first choose $L_{\max}$ and set $L_{\min} = (1-\mathrm{skew})L_{\max}$. We then assign each query tile a key/value length sampled between $L_{\min}$ and $L_{\max}$, and randomly select that many key/value blocks for the tile. This method lets us control the degree of load imbalance while keeping the sparsity fixed.

\begin{figure}[!htb]
\centering
\includegraphics[width=\linewidth]{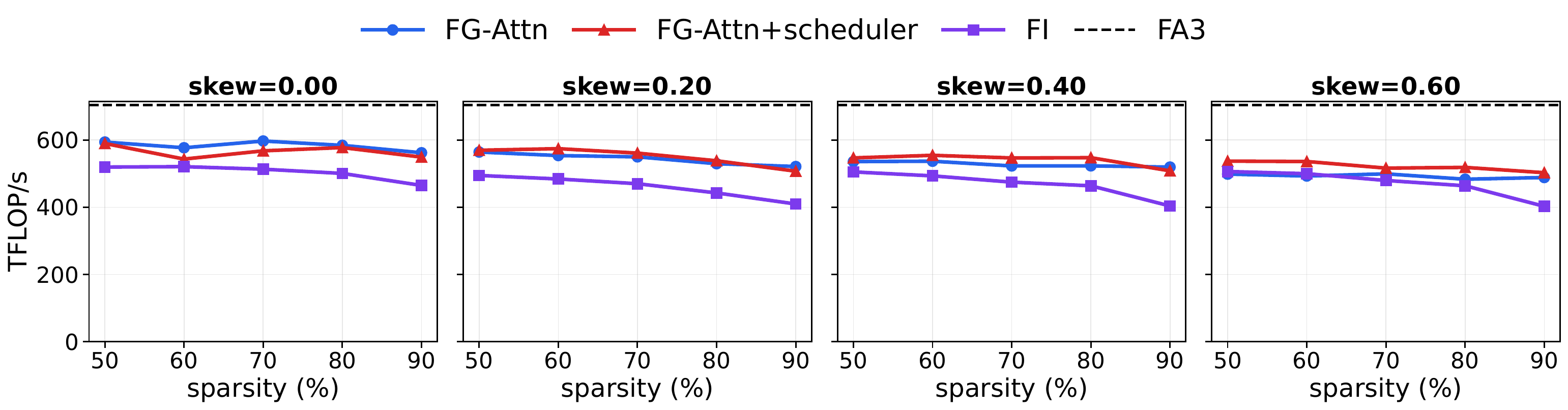}
\caption{Achieved compute throughput of \X and \X (no scheduler) compared to dense flash attention at different sparsities, at varying levels of skew in key/value sequence lengths. }
\label{fig:attn_kernel_analysis}
\end{figure}

Fig.~\ref{fig:attn_kernel_analysis} shows the compute throughput (TFLOPs/s) of FA3, \X with the persistent work scheduler (\X), and \X without the scheduler (\X (no scheduler) ), across different sparsity levels and degrees of skew. We chose a block size $128\times 32$ for this experiment. From these experiments, we make three observations.
First, \X achieves throughput within $18\%$ of dense FA3. This is significant because \X executes fine-grained sparse attention, which requires irregular key/value gathers, sparse-mask processing, and dynamic work scheduling, whereas FA3 operates on dense, contiguous tiles. Despite these additional overheads, \X preserves most of the compute efficiency of dense attention while skipping masked key/value blocks. This shows that \X better converts sparsity into high tensor core utilization, rather than incurring performance losses due to scheduler overheads or irregular execution.

Second, the persistent work scheduler improves utilization by assigning output tiles to SMs dynamically based on their estimated work. As a result, \X improves throughput by up to $11\%$ over \X (no scheduler), which statically assigns work to SMs and can leave some SMs idle when different query tiles attend to different numbers of key/value tiles.
Third, the throughput gap between \X and \X (no scheduler) widens as skew increases. At higher skew, the variation in key/value lengths across query tiles becomes larger, making load imbalance more severe. The persistent work scheduler mitigates this imbalance by prioritizing tiles with larger estimated workloads, reducing tail effects, and improving SM utilization.
From these experiments, we conclude that \X provides a highly efficient kernel for fine-grained sparse attention, that operates efficiently at different levels of skew.

\vspace{-0.1cm}
\subsection{Evaluating End-to-End Video Generation with Sparse Attention}
\label{sec:main_result}
\vspace{-0.1cm}
%We now compare the end-to-end video generation time of \X. 
% The corresponding results for \X+consecutive merge are given in the Appendix. 
Table~\ref{tab:video_generation_comparison} summarizes video generation time, attention FLOPs normalized to dense attention, and video quality relative to dense attention, respectively. We make the following observations.
First, \X speeds up video generation across all evaluated video models. Compared to the dense FA3 baseline, \X-16, \X-32, and \X-64 achieves speedups of up to $1.26\times$, $1.40\times$, and $1.40\times$, respectively, with average speedups of $1.11\times$, $1.18\times$, and $1.19\times$. These speedups come from reducing the amount of attention computation. %On average, \X-16 reduces attention FLOPs to $0.37\times$ of dense attention.
Second, fine-grained sparse attention exposes more opportunities to skip unnecessary attention computation. The \X-16 configuration has the lowest attention FLOPs on average, reducing attention FLOPs to $0.37\times$ of dense attention, compared to $0.40\times$ for \X-32 and $0.44\times$ for \X-64. This is because smaller sparse blocks can more precisely remove irrelevant query--key regions, whereas coarser blocks must retain a larger region even when only part of the block is useful. At the same time, finer granularity preserves video quality better: compared to FI+SVG2, \X-16 improves PSNR by $0.86$ on average. %Coarse-grain sparse attention incurs a higher latency in mask computation time.
Third, compared to FI+SVG2, \X achieves higher end-to-end speedups while maintaining or improving video quality. \X-16, \X-32, and \X-64 is faster by up to $2.06\times$, $2.20\times$, and $2.20\times$, respectively, with average speedups of $1.42\times$, $1.48\times$, and $1.50\times$. Although FI+SVG2 sometimes reduces attention FLOPs more aggressively, it incurs substantial overhead from sparse-mask processing, query clustering/reordering, and FI planning. These overheads are especially visible for smaller-context models such as LTX-2, where FI+SVG2 is slower than dense FA3 despite reducing attention FLOPs to $0.44\times$ of dense attention. In contrast, \X more effectively translates attention FLOP reduction into end-to-end generation speedup.
Fourth, SpargeAttention, SVG, and Radial Attention rely on coarse block-sparse attention, which can lead to larger quality loss because useful query--key scores may be skipped together with unimportant scores in the same block. In contrast, \X enables better video quality using fine-grained masks that more precisely retain useful attention regions. % Moreover, these mask strategies are compatible with \X.
Overall, \X improves video generation efficiency by combining fine-grained sparse attention with low-overhead sparse execution. Finer block sizes expose more sparsity and better preserve quality, while larger block sizes can provide higher end-to-end speedups due to lower mask computation overheads. 
% Across models, \X consistently improves over dense FA3 and avoids the clustering overheads of FI+SVG2.

% \input{tables/hyvideo_table}

% \input{tables/wan14b_table}

% \input{tables/wan1.3b_table}

% \input{tables/ltx2_table}

\input{tables/combined}

% \subsection{Ablations}

\begin{figure}[!htb]
\vspace{-0.45cm}
\centering
\begin{subfigure}{0.60\textwidth}
    \centering
    \includegraphics[width=\linewidth]{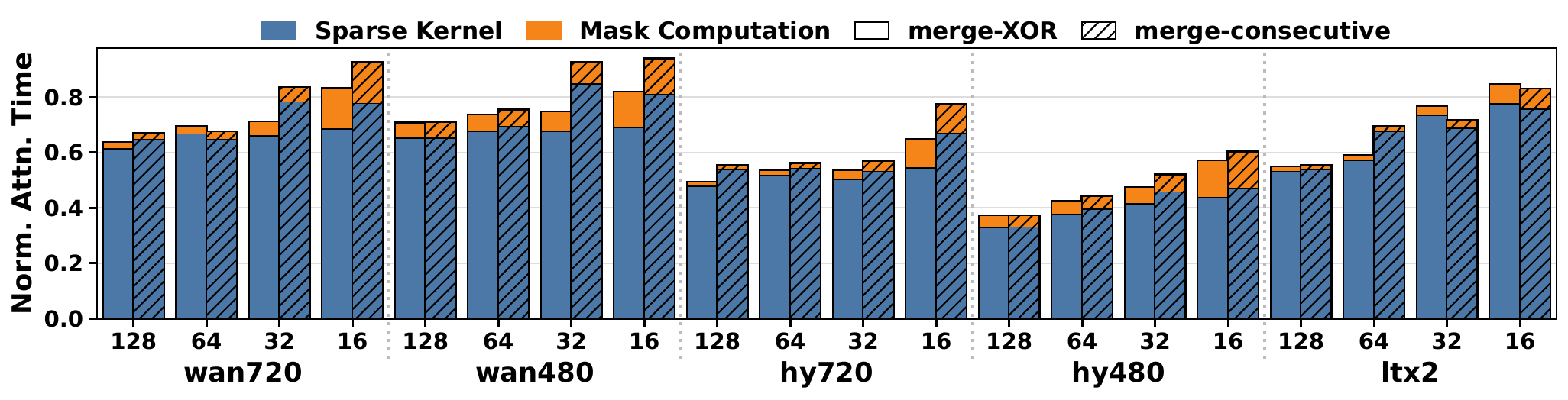}
    \caption{Attention time breakdown at different block sizes.}
\label{fig:attention_breakdown}
\end{subfigure}~
\begin{subfigure}{0.39\textwidth}
    \centering
    \includegraphics[width=\linewidth]{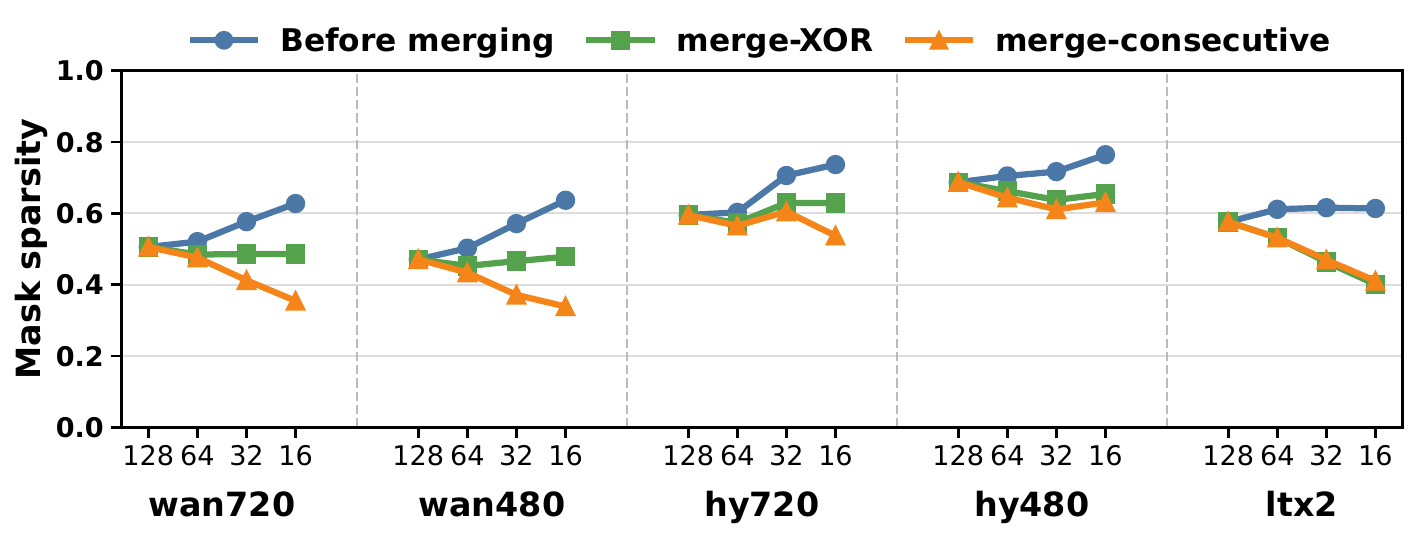}
    \caption{Sparsity at different block sizes.}
    \label{fig:density_bars}
\end{subfigure}
\caption{Sparsity and attention computation breakdown at different block sizes.}
\vspace{-0.45cm}
\end{figure}

% \vspace{-0.3cm}
\label{sec:ablation}
\textbf{Ablation: Attention computation cost vs. block size.}
Fig.~\ref{fig:attention_breakdown} shows the attention and mask computation time breakdown normalized to FA3. We observe that \X is faster than dense attention in all evaluated settings, with total attention time ranging from $0.37\times$ to $0.94\times$ of dense attention, showing that \X can leverage sparsity to speed up attention.
The mask computation takes a small amount of time on average ($0.06\times$ of FA3), showing that the sparse mask is cheap to compute.
However, the mask overhead increases at finer block sizes, from $0.03\times$ at $128\times128$ to $0.12\times$ at $16\times16$. Merge-XOR outperforms consecutive merging because it reduces redundant key/value loads.
Overall, \X reduces attention time while keeping mask overhead small.

\textbf{Ablation: Sparsity on merging query tiles.}
Fig.~\ref{fig:density_bars} shows the block sparsity before merging query tiles, and after merging to 128 queries using minimum XOR distance (\X+merge-XOR), and merging consecutive elements (\X+merge-consecutive).
First, we observe that smaller blocks expose more sparsity before merging: the average sparsity increases from $0.57$ at $128\times128$ to $0.68$ at $16\times16$. Both merge-XOR and merge-consecutive decrease the effective sparsity. Third, merge-XOR preserves more sparsity than consecutive merging, especially at finer granularities. At $16\times16$, merge-XOR merging gives an average density of $0.47$, compared to $0.55$ for merge-consecutive.
Overall, finer blocks expose more sparsity, but query merging can reduce the realized benefit. merge-XOR reduces this loss and makes fine-grained sparse attention more efficient.

\vspace{-0.45cm}

% \begin{figure}[!htb]
%     \centering
%     \includegraphics[width=\linewidth]{result_figs/ltx2.pdf}
%     \caption{for various and the corresponding video produced at various block sizes. Using larger block sizes will degrade quality.}
%     \label{fig:ltx2_images}
% \end{figure}

%% file: tables/combined.tex
\begin{table}[!htb]
\vspace{-0.5cm}
\centering
\small
\setlength{\tabcolsep}{4.5pt}

% ---------------- HunyuanVideo ----------------
\begin{tabular}{|l|cccc|cccc|}
\hline
& \multicolumn{4}{c|}{\textbf{HunyuanVideo 480p, 129 Frames}}
& \multicolumn{4}{c|}{\textbf{HunyuanVideo 720p, 129 Frames}} \\
\cline{2-9}
\textbf{Method}
& Time & Attn. FLOPs & PSNR & SSIM
& Time & Attn. FLOPs & PSNR & SSIM \\
\hline
SVG        & -    & 0.3X  & 20.20 & 0.79 & -     & 0.3X  & 21.19 & 0.76 \\
Sparge     & -    & -     & $<20$ & 0.79 & -     & -     & $<20$ & $<0.7$ \\
Radial     & -    & 0.37X & 21.50 & 0.80 & -     & 0.37X & 22.33 & 0.77 \\
\hline
FA3        & 3:06 & 1X    & inf.  & 1    & 17:21 & 1X    & inf.  & 1 \\
FI+SVG2    & 3:00 & 0.29X & \textbf{23.63} & \textbf{0.83} & OOM   & OOM   & OOM   & OOM \\
\hline
FG-Attn-16 & 2:55 & \textbf{0.31X} & 23.24 & \textbf{0.83} & 13:49 & \textbf{0.27X} & \textbf{22.40} & \textbf{0.81} \\
FG-Attn-32 & 2:43 & 0.32X & 22.80 & 0.83 & 12:26 & 0.28X & 22.02 & 0.76 \\
FG-Attn-64 & \textbf{2:40} & 0.34X & 22.21 & 0.81 & \textbf{12:26} & 0.30X & 22.21 & 0.74 \\
\hline
\end{tabular}

% \vspace{0.25cm}
% \vspace{-0.05cm}

% ---------------- Wan ----------------
\begin{tabular}{|l|cccc|cccc|}
\hline
& \multicolumn{4}{c|}{\textbf{Wan 2.1 14B 480p, 81 Frames}}
& \multicolumn{4}{c|}{\textbf{Wan 2.1 14B 720p, 81 Frames}} \\
\cline{2-9}
\textbf{Method}
& Time & Attn. FLOPs & PSNR & SSIM
& Time & Attn. FLOPs & PSNR & SSIM \\
\hline
SVG        & -    & 0.3X  & 25.59 & 0.84 & -     & 0.3X  & 22.39 & 0.80 \\
Sparge     & -    & -     & 24.11 & 0.79 & -     & -     & 23.51 & 0.78 \\
Radial     & -    & 0.37X & 25.01 & 0.80 & -     & 0.37X & 24.31 & 0.78 \\
\hline
FA3        & 4:50 & 1X    & inf.  & 1    & 18:34 & 1X    & inf.  & 1 \\
FI+SVG2    & 5:25 & 0.31X & 26.47 & 0.87 & 17:31 & 0.37X & 25.36 & \textbf{0.84} \\
\hline
FG-Attn-16 & 4:46          & \textbf{0.34X} & \textbf{26.50} & \textbf{0.87} & 17:10 & \textbf{0.34X} & \textbf{25.39} & \textbf{0.84} \\
FG-Attn-32 & 4:26          & 0.37X          & 26.22 & 0.87 & 16:36 & 0.40X & 24.02 & 0.80 \\
FG-Attn-64 & \textbf{4:20} & 0.40X          & 25.82 & 0.86 & \textbf{15:37} & 0.45X & 23.24 & 0.78 \\
\hline
\end{tabular}

% \vspace{-0.05cm}
% \vspace{0.25cm}

% ---------------- LTX2 ----------------
\begin{tabular}{|l|cccc|cccc|}
\hline
& \multicolumn{4}{c|}{\textbf{LTX2 512x768, 121 Frames}}
& \multicolumn{4}{c|}{\textbf{LTX2 1024x1536, 121 Frames}} \\
\cline{2-9}
\textbf{Method}
& Time & Attn. FLOPs & PSNR & SSIM
& Time & Attn. FLOPs & PSNR & SSIM \\
\hline
FA3           & 0:36      &  1X           & inf.     &  1.0      & 3:08  &   1X         & inf.  &  1.0  \\
FI+SVG2       & 1:00 & 0.44X & 27.03 & 0.88 & 4:55 & 0.43X & 25.36 & 0.88 \\
\hline
FG-Attn-16 & 0:33           & \textbf{0.48X} & \textbf{28.27} & \textbf{0.90} & 2:51            & \textbf{0.48X} & \textbf{28.73} & \textbf{0.93} \\
FG-Attn-32 & \textbf{0:31}  & 0.52X          & 27.35          & 0.90          & 2:45            & 0.51X          & 28.29          & 0.92 \\
FG-Attn-64 & \textbf{0:31}  & 0.58X          & 26.77          & 0.89          & \textbf{2:45}   & 0.59X          & 27.64          & 0.91 \\
\hline
\end{tabular}

\caption{Comparing video generation times, attention FLOPs, and generation quality.}
\label{tab:video_generation_comparison}
\vspace{-0.65cm}
\end{table}

%% file: sections/conclusion.tex
\section{Conclusion}
\label{sec:conclusion}
\vspace{-0.3cm}
We introduced \textbf{\X}, a fine-grained sparse attention mechanism that skips query--key computations at fine block granularities without incurring the overheads of redundant loads/computation and irregular memory access on modern GPUs. \X provides a practical training-free substrate to exploit fine-grain sparsity that can be seamlessly integrated into state-of-the-art video and image diffusion transformers. \X achieves significant end-to-end speedups with negligible quality loss, thus subsuming existing block-sparse attention mechanisms. %These results demonstrate that fine-grained sparsity can be realized efficiently, providing a scalable alternative that subsumes block-sparse attention in DiTs. 

%% file: sections/appendix.tex
\section{GPU Architecture Overview}
\label{sec:background_gpu}
\vspace{-0.2cm}

% Fig.~\ref{fig:gpuarch} provides a high-level overview of modern GPU architecture. Tensor data is first transferred from high-bandwidth memory (HBM) into the shared memory of streaming multiprocessors (SMs), from which threads schedule computations on the tensor cores

\begin{figure}[!htb]
    \centering
    \includegraphics[width=0.4\linewidth]{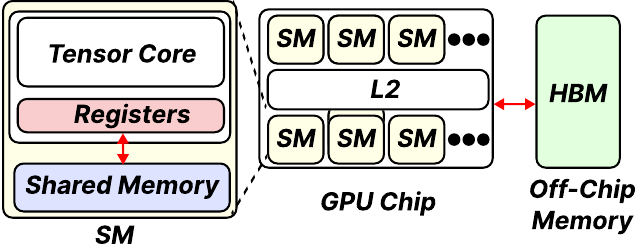}
    \caption{High-level view of modern GPU architecture.}
    \label{fig:gpuarch}
    \vspace{-0.3cm}
\end{figure}

\paragraph{Compute hierarchy.}
Fig.~\ref{fig:gpuarch} provides a high-level overview of modern GPU architecture.
A GPU consists of many \emph{streaming multiprocessors} (SMs). Each SM contains its own register file, on-chip shared memory (or scratchpad memory), and a set of \emph{tensor cores} for matrix multiplication. Each SM executes many \emph{threads}. 
% Threads are organized into \emph{thread blocks}, which are scheduled onto SMs.
Threads are grouped at three levels: 32 threads form a \emph{warp} that executes the same instruction in lockstep. A fixed number of warps form a \emph{thread block} that is scheduled onto a single SM. 
On Hopper GPUs, four warps (128 threads) form a \emph{warp-group}, the unit at which a tensor core matrix multiplication instruction is issued.

A \emph{kernel} is a function that the programmer launches with a grid of thread blocks; each block runs on one SM. The programmer decides how many threads per block, how much shared memory each block uses, and how work is partitioned across blocks. 
%this last choice is the focus of our scheduler in Section~\ref{sec:scheduling}.

\paragraph{Memory hierarchy.}
GPU memory consists of broadly three types: \emph{High-bandwidth memory} (HBM) is the off-chip RAM (e.g., 80~GB on an H100) that holds the model's weights, activations, and KV cache; it has the highest capacity but also the highest latency, on the order of hundreds of cycles. Each SM has \emph{shared memory} (typically tens to a few hundred KB), an on-chip scratchpad that all threads in a block can access in a few cycles. Each thread additionally has private \emph{registers}, the fastest storage but limited in number. Achieving high performance on a GPU is largely about (a) keeping data resident in shared memory and registers, and (b) overlapping the unavoidable HBM transfers with compute.

\paragraph{Tensor cores.}
Tensor cores are specialized matrix-multiply units that operate on small tiles (e.g., $64 \times 64 \times 16$ on Hopper) at very high throughput. To use them, a kernel must (i) stage operand tiles into shared memory or registers, (ii) issue a tensor-core instruction (\texttt{wgmma} on Hopper), and (iii) wait for the result. A modern attention kernel spends most of its time issuing tensor-core matmuls for $QK^\top$ and the subsequent multiplication by values. 
% The surrounding logic exists primarily to feed these matmuls without stalling them.

% Fig.~\ref{fig:gpuarch} provides a high-level overview of modern GPU architecture. A GPU consists of multiple streaming multiprocessors (SMs), each of which executes many \emph{threads}. Threads are organized into \emph{thread blocks}, which are scheduled onto SMs. Within a thread block, threads are further organized into \emph{warps}, groups of threads that execute instructions together. Modern attention kernels use these threads to move tensor data from high-bandwidth memory (HBM) into the SM's shared memory and to schedule matrix operations on tensor cores.

\begin{figure}[!htb]
    \centering
    \includegraphics[width=0.5\linewidth]{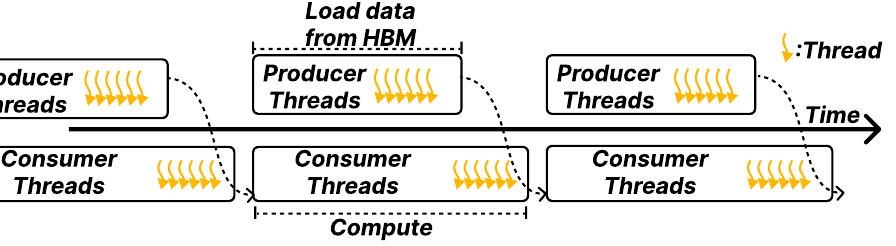}
    \caption{Pipelined execution of producer and consumer threads: Data is prefetched by the producer while the consumer threads are doing computation.}
    \label{fig:pipelining_wg}
    \vspace{-0.3cm}
\end{figure}

\paragraph{Warp specialization and producer--consumer pipelining.}
Because HBM latency is so much larger than tensor-core throughput, naively loading a tile and then computing on it leaves the tensor cores idle for most of every iteration. Modern attention kernels hide this latency through \emph{warp specialization}: within a thread block, some threads are designated \emph{producers} and others \emph{consumers}, as shown in Fig.~\ref{fig:pipelining_wg}. Producer threads issue asynchronous loads (using \texttt{cp.async} or, on Hopper, the Tensor Memory Accelerator) that copy tiles of $K$ and $V$ from global to shared memory. Consumer warp-groups wait on a barrier until a tile is ready, multiply them with a group of queries using the tensor core matrix multiply operation. Upon completion, they signal the producer threads to load the next key-value pair into shared memory. 
With multiple buffer slots in shared memory, the producer can run several iterations ahead of the consumer, so that by the time the consumer finishes one matmul, the next tile is already resident on-chip. This is the software pipeline depicted in Fig.~\ref{fig:pipelining_wg}. Producer threads issue load operations to move data from HBM into shared memory, while consumer threads schedule computations on the tensor cores using the fetched data. By overlapping these operations, GPU resources remain efficiently utilized.

% Threads within a GPU thread block are divided into producers and consumers, as shown in Fig.~\ref{fig:pipelining_wg}. Producer threads issue load operations to move data from HBM into shared memory, while consumer threads schedule computations on the tensor cores using the fetched data. By overlapping these operations, GPU resources remain efficiently utilized. This pipelined execution is illustrated in Fig.~\ref{fig:pipelining_wg}.

\section{Persistent Work Scheduler}
\label{sec:load_balance}
\vspace{-0.2cm}
% In Flash Attention~\cite{fa}, each attention head maps a group of query tokens to a single thread block to produce a set of output tokens. In \X, query groups (of variable sizes) iterate over a variable number of key--value pairs to compute attention scores, which can cause load imbalance: some thread blocks sit idle while others process longer key--value lists.

% To mitigate this, \X employs a persistent work-stealing scheduler, depicted in Fig.~\ref{fig:scheduling}. In Fig.~\ref{fig:scheduling}, each 128-query block is processed by a thread block, shown in green in Fig.~\ref{fig:work_per_output}. This query block is going to attend to a different number of keys/values, indicated by the grey bars .

% Before scheduling, the amount of work to compute each output tile is set to the number of keys/values the query group it is going to attend to (Fig.~\ref{fig:work_per_output}) \todo{grammar issues}. %KV lengths .
% The scheduler then follows a longest-processing-time-first (LPT) policy, prioritizing work items whose estimated runtime is longest. Specifically, this corresponds to those requiring iteration over the most key-value pairs. We implement this technique in the kernel itself. A persistent-thread work-stealing algorithm on the GPU then dispatches these highest-cost blocks first, keeping utilization high across thread blocks.

% \todo{what is a persistent work-stealing scheduler}

\begin{figure}[!htb]
\begin{subfigure}{0.38\textwidth}
    \centering
    \includegraphics[width=\linewidth]{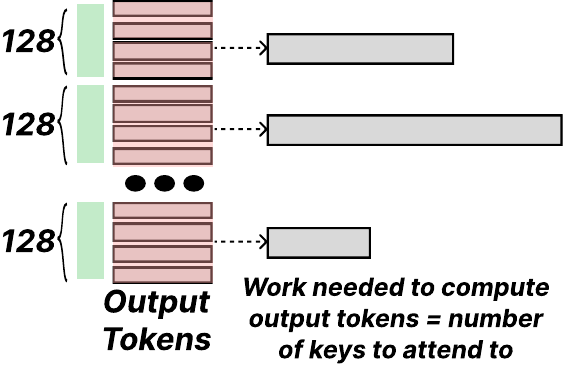}
    \caption{Number of keys/values (KVs) loaded for each query block is first computed.}
    \label{fig:work_per_output}
\end{subfigure}~
\begin{subfigure}{0.59\textwidth}
    \centering
    \includegraphics[width=0.8\linewidth]{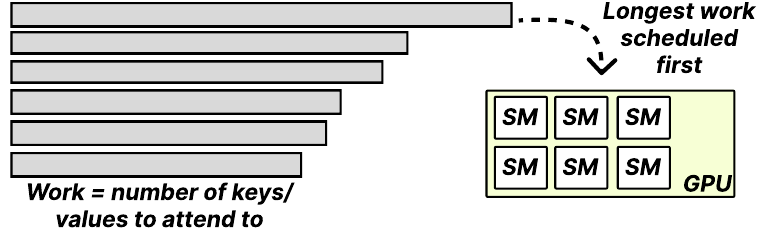}
    \caption{The query block requiring the longest processing time (proportional to the number of KVs to be processed) is scheduled first.}
    \label{fig:work_scheduling}
\end{subfigure}
\caption{Scheduling work across output tiles: the cost of each tile is determined by the number of key--value pairs its query block attends to.}
\label{fig:scheduling}
% \vspace{-0.3cm}
\end{figure}

In Flash Attention~\cite{fa}, each attention head assigns a group of 128 query tokens to a single thread block, which produces the corresponding output tokens. In \X, however, query groups (of size 128 after merging) attend to a variable number of key--value pairs, which causes load imbalance: some thread blocks finish quickly and sit idle while others grind through much longer key--value lists.
To address this, \X uses a persistent work-stealing scheduler (Fig.~\ref{fig:scheduling}). Each 128-query block (shown in green in Fig.~\ref{fig:work_per_output}) is processed by one thread block and attends to a different number of key--value pairs, indicated by the grey bars. Before dispatch, we estimate the cost of each output tile as the length of its associated key--value list (Fig.~\ref{fig:work_per_output}).

The scheduler then applies a longest-processing-time-first (LPT) policy, prioritizing the tiles with the largest key--value lists, since these have the longest expected runtime. Similar LPT-based scheduling is used in FlashInfer~\cite{flashinfer}, FA2~\cite{fa2}, and FA3~\cite{fa3}; FlashInfer, for instance, builds the schedule on the CPU using a min-heap. We go further by making the scheduler persistent and tailoring it to variable block-sparse attention, which eliminates the kernel-launch and tile-assignment overheads that would otherwise dominate at our granularity. Concretely, each thread block fetches the highest-cost remaining tile directly from a GPU-side work queue, processes it, and then loops back for the next one, so all thread blocks stay busy until the head is complete.